\pdfoutput=1

\documentclass[11pt]{article}

\usepackage[]{ACL2023}

\usepackage{times}
\usepackage{latexsym}

\usepackage[T1]{fontenc}

\usepackage[utf8]{inputenc}

\usepackage{microtype}

\usepackage{inconsolata}

\usepackage[utf8]{inputenc}
\usepackage[english]{babel}
\usepackage{listings}  
\usepackage{url}
\usepackage{amsmath}
\usepackage{paralist}  
\usepackage{color}
\usepackage{float}
\usepackage{amsfonts}
\usepackage{amssymb}
\usepackage{amsmath}
\usepackage{bm}
\usepackage{booktabs}
\usepackage{tabularx}
\usepackage{hyperref}
\usepackage{graphicx}
\graphicspath{ {./images/} }
\title{Evaluating Embedding Frameworks for Scientific Domain}

\author{
  Nouman Ahmed \\
  Iris.ai/University of Oxford  \\\And
  Ronin Wu \\
  Iris.ai/QunaSys \\\And
  Victor Botev \\
  Iris.ai
  } 
\begin{document}
\maketitle
\begin{abstract}
Finding an optimal word representation algorithm is particularly important in terms of domain specific data, as the same word can have different meanings and hence, different representations depending on the domain and context. While Generative AI and transformer architecture does a great job at generating contextualized embeddings for any given work, they are quite time and compute extensive, especially if we were to pre-train such a model from scratch. In this work, we focus on the scientific domain and finding the optimal word representation algorithm along with the tokenization method that could be used to represent words in the scientific domain. The goal of this research is two fold: 1) finding the optimal word representation and tokenization methods that can be used in downstream scientific domain NLP tasks, and 2) building a comprehensive evaluation suite that could be used to evaluate various word representation and tokenization algorithms (even as new ones are introduced) in the scientific domain. To this end, we build an evaluation suite consisting of several downstream tasks and relevant datasets for each task. Furthermore, we use the constructed evaluation suite to test various word representation and tokenization algorithms.
\end{abstract}

\section{Introduction}
\label{subsec:intro}
Effective word representation is crucial in Natural Language Processing (NLP), significantly impacting tasks such as language understanding, text generation, and sentiment analysis. Optimal representation enables NLP models to grasp contextual and semantic relationships among words, addressing challenges like handling out-of-vocabulary words and performing word similarity and analogies. This importance extends to various applications, including machine translation, multilingual NLP, and pre-training for transfer learning.

In the realm of word representation, two key areas of research, tokenization, and word embeddings, play pivotal roles. Tokenization involves splitting raw text into smaller sub-texts, a fundamental step in NLP tasks. However, challenges arise in languages without clear word boundaries, where tokenization demands careful consideration. On the other hand, various methods exist for representing words, ranging from traditional approaches like One-Hot Encoding and TF-IDF to more sophisticated techniques such as Distributional Models and Word Embeddings.

Despite the diversity of representation methods, Word Embeddings have become the de-facto standard in NLP due to their ability to capture semantic and contextual information efficiently. These embeddings, often evaluated based on lexical semantics tasks, are critical building blocks for downstream NLP tasks. However, evaluating their performance on specific domains, such as scientific literature, remains an open challenge.

This paper focuses on evaluating tokenization and word representation algorithms within the scientific domain. While general-purpose data often suffices for training word embeddings, scientific texts introduce unique challenges, demanding specialized handling. Recent models like SciBERT \cite{beltagy-etal-2019-scibert}, trained specifically on scientific domain data, showcase the necessity for domain-specific embeddings. What is interesting here is that SciBERT creates a new vocabulary specific to scientific domain, which only has a overlap of 42\% with the general vocabulary of BERT \cite{Devlin2019BERTPO}, showing that scientific data is quite different from general purpose data and should be handled as such. But training these transformer models comes at the expense of time and compute, with SciBERT requiring a TPU with 8 cores and 1 week to be trained from scratch. In comparison, simpler (static) embeddings like Word2Vec tend to use fewer resources and can generate relatively good embeddings, especially in the case of lack of context.

\textbf{Research Objectives: }The primary goal of this research is to address the gap in understanding how different tokenization and word representation methods perform in scientific NLP tasks. The goal of the research, thus, contains of the following objectives:

\begin{enumerate}
    \item Building a comprehensive evaluation suite consisting of various downstream NLP Tasks related to scientific domain, covering both intrinsic and extrinsic evaluation
    \item Testing various word embedding and tokenization algorithms against this evaluation suite
    \item Comparing low-resource word embedding models, such as Word2Vec, to computationally extensive Transformer-based models, especially when trained on large domain-specific data
    \item Measure the strength of generalization by including some general-purpose domain datasets into the evaluation suite
\end{enumerate}

By fulfilling these objectives, we aim to provide insights into the domain-specific challenges of word representation in scientific NLP, contributing towards the development of more effective and adaptable models.

\section{Related Work}
The related work section of this paper is divided into the three areas of research, namely, Tokenization, Word Embeddings, and Evaluation of Word Representation algorithms.

\subsection*{Tokenization}
In the realm of Natural Language Processing (NLP), tokenization plays a critical role, especially in handling the vast and dynamic vocabulary of languages. Byte Pair Encoding (BPE) \cite{Sennrich2015NeuralMT} is a pioneering sub-word tokenizer used in transformer models, including GPT and BERT, demonstrating its relevance in state-of-the-art NLP tasks. However, BPE faces challenges such as handling out-of-vocabulary characters and producing sub-optimal tokenizations \cite{Bostrom2020BytePE}. WordPiece \cite{Wu2016GooglesNM}, employed by models like DistilBERT and MobileBERT, refines the BPE approach by considering the likelihood of sub-word pairs, offering faster tokenization with less memory usage. Unigram Tokenizer \cite{Kudo2018SubwordRI} takes a distinct approach, starting with a large vocabulary and iteratively removing tokens to achieve the desired size, resulting in more logical sub-strings \cite{Bostrom2020BytePE}. SentencePiece \cite{Kudo2018SentencePieceAS} addresses the limitation of assuming space-separated words by treating the input as a raw stream, making it suitable for languages with concatenated words. 

\subsection*{Word Embeddings}

Word embeddings \cite{Firth1957ASO} serve as numerical representations of words or tokens, capturing both semantic and contextual meaning. Various algorithms contribute to generating these embeddings, with Word2Vec \cite{Mikolov2013EfficientEO} and GloVe \cite{Pennington2014GloVeGV} being influential static models. FastText \cite{Bojanowski2016EnrichingWV} extends Word2Vec by incorporating n-grams to address rare or out-of-vocabulary (OOV) words more effectively. Contextual embedding models like Context2Vec \cite{Melamud2016context2vecLG} and CoVE \cite{McCann2017LearnedIT} leverage bidirectional LSTMs, bridging the contextual information gap. Newer approaches such as HiCE \cite{Hu2019FewShotRL} introduce few-shot learning to enhance representation. Additionally, multi-level OOV handling strategies \cite{Lochter2022MultilevelOW} and DL language models like GPT \cite{brown2020language} and BERT \cite{Devlin2019BERTPO} focus on generating dynamic embeddings using data such as the token's position and context. 

\subsection*{Evaluation}

Word embedding models demand thorough evaluation, and researchers have conducted studies focusing on both intrinsic and extrinsic aspects. Intrinsic evaluation assesses the quality of word embeddings, often measuring word similarity through synonyms and antonyms, where as extrinsic evaluation focuses on the embeddings' real-world applicability in downstream tasks such as classification. \cite{Lochter2020DeepLM} explored Deep Learning (DL) models' intrinsic and extrinsic performance on out-of-vocabulary (OOV) tokens using the Chimera dataset \footnote{https://github.com/UCLA-DM/HiCE/tree/master/data/chimeras} \cite{Lazaridou2013CompositionallyDR}, by means of tasks such as word similarity and text categorization. \cite{Rogers2018WhatsIY} conducted intrinsic tasks, such as word similarity and analogy, and extrinsic tasks  including named entity recognition and natural language inference, offering a comprehensive evaluation with task-specific metrics and exploring correlations among various linguistic factors. 
\section{Evaluation Framework}
To fulfil our research objectives, as outlined in section \ref{subsec:intro}, mainly on finding the optimal tokenization and word representation models for scientific domain, we need to construct an evaluation suite. This means selecting the various individual downstream tasks and their relevant datasets, in order to evaluate our tokenization methods and embedding models. 
Below is a small summary of the tasks that we focus on, grouped by the type of evaluation:

\subsection{Tasks}
\subsubsection{Intrinsic Evaluation}
    \begin{itemize}
        \item Word and Sentence Similarity: This will help us making sure that the different words and sentences used for the same context are embedded quite similarly and not differ by a lot. This will be particularly helpful for some systems built on top of these embeddings such as a recommendation system for suggesting relevant articles.
    \end{itemize}

  \subsubsection{Extrinsic Evaluation}
    \begin{itemize}
        \item Named Entity Recognition (NER): We focus on this because extracting entities from scientific domains is useful in downstream tasks like summarization or extracting quantities of certain chemicals (entities).
        \item Classification: This will help us classify scientific reports (positive or negative) or papers (as being against a particular stance or something).
    \end{itemize}

\subsection{Datasets}
Below is a brief description of the datasets selected for each task. For a detailed description of the tasks and the relevant datasets, see Appendix \ref{sec:eval_framework_appendix}

\subsubsection{Word Similarity}
    \begin{itemize}
        \item UMNSRS: University of Minnesota Semantic Relatedness Standard (UMNSRS) \cite{Pakhomov2018SemanticRA} is a collection of concept pairs that are manually annotated for both similarity and relatedness using a continuous response scale
    \end{itemize}

\subsubsection{Sentence Similarity}
    \begin{itemize}
        \item SemEval-2017: SemEval is an international workshop and a shared task \cite{cer-etal-2017-semeval} on Semantic Evaluation held every year which aims to measure semantic textual similarity (STS) given a dataset. The dataset comprises a selection of the English datasets used in the STS tasks organized in the context of SemEval between 2012 and 2017
        \item Clinical STS 2018 | 2019: This was introduced as a shared task in SemEval, in an attempt to improve the systems for semantic textual similarity specific to medical data \cite{Wang2018MedSTSAR}. Two annotators, having vast experience of many years in clinical domain, annotated the sentences manually, with weighted Cohen's Kappa of 0.67.
    \end{itemize}
\subsubsection{Named Entity Recognition}
    \begin{itemize}
        \item CoNLL-2003:  The dataset \cite{Sang2003IntroductionTT} was introduced as a shared task. The NER part of the shared task concentrates on four types of named entities: persons, locations, organizations, and miscellaneous entities. 
        \item CHEMDNER: CHEMDNER \cite{Krallinger2015TheCC} is chosen as another NER dataset for our evaluation suite as it covers the scientific domain, specifically chemistry. CHEMDNER is a collection 10,000 PubMED abstracts, containing a total of 84,355 chemical entity mentions labeled manually by expert chemistry literature curators. 
        \item SciERC: SciERC \cite{Luan2018MultiTaskIO} was introduced as a multi-task setup in order to identify entities, relations, and co-reference clusters in scientific articles. Here, we mainly focus on identifying entities. It includes scientific entities for 500 AI-community specific scientific abstracts collected from the Semantic Scholar Corpus \cite{Lo2020S2ORCTS}. 
    \end{itemize}
\subsubsection{Document Classification}
    \begin{itemize}
        \item Reuters 12: Reuters \footnote{\url{https://martin-thoma.com/nlp-reuters/}} is a benchmark dataset for document classification. Originally, it is a multi-class and multi-label (each document can have more than one class) dataset. For our purposes, we selected a subset of the dataset. The subset of the dataset contained 12 (out of original 90) classes and is a single label, multi-class dataset
        \item Biochem 8: Biochem dataset is a document classification dataset containing biomedical documents classified into 8 different classes such as abortion, cancer, etc. The dataset is a multi-class, single label dataset. This is a confidential dataset from a company's client \footnote{\url{https:iris.ai}}
    \end{itemize}

\section{Methodology}

\subsection{Embedding Models}
In this section, we will go through the word representation models that we will be working with in our evaluation suite. We are focusing on three major type of word representation algorithms:
\begin{itemize}
    \item \textbf{Word2Vec: }As Word2Vec is one of the early algorithms used for word representation, and is still used in a lot of NLP tasks, it seems logical to start with this. And mainly because it is a low-resource word representation model, in comparison to the current state-of-the-art i.e, Generative AI and Transformer based models. A Word2Vec model can be trained without a GPU and generally within minutes as compared to requiring specific GPU/TPUs and days to train a transformer model. We train the model on two different datasets, one generic and one scientific domain specific (more on it later). We also train the embedding model for different dimensions and different underlying mechanisms as well, i.e., Cbow and Skipgram (explained below).
    \item \textbf{FastText: } Building on Word2Vec, we move to FastText as it incorporates the n-gram knowledge into the word representation model, in addition to also being able to construct an embedding for a out-of-vocabulary (OOV) word using these n-grams. Same as with word2vec, we train it on different datasets, dimensions and underlying mechanisms.
    \item \textbf{Transformer Based: } Finally, we experiment with transformer based models. The reason being that these models power some of the most used systems nowadays and usually out-perform simple word-representation models, albeit at more computational cost as mentioned above. As these models generate dynamic embeddings, these were not fine-tuned on any dataset (as that requires considerable compute), rather uses the context in the specific task to generate an embedding for the model. Note that these models have some kind of tokenization (generally BPE) already built-in.

\end{itemize}

Both Word2Vec and FastText embeddings can be trained using two different underlying mechanisms, mainly Continuous Bag of Words (CBOW) or Skip-gram.

\begin{itemize}
    \item \textbf{CBoW: }The cbow model predicts the target word according to its context. The context is represented as a bag of the words contained in a fixed size window around the target word.
    \item \textbf{Skipgram: }The skipgram model learns to predict a target word thanks to a nearby word.
\end{itemize}

\subsection{Tokenization Methods}

\begin{itemize}
    \item \textbf{Byte Pair Encoding: }Introduced initially as a text compression algorithm, Byte Pair Encoding (BPE) has become a prevalent subword tokenizer in modern NLP. The algorithm begins by considering individual characters in the corpus, constructing a base vocabulary that forms the starting point for subword segmentation. BPE iteratively merges the most frequent character pairs, updating the vocabulary with each merge.
    \item \textbf{WordPiece:} Similar to BPE, WordPiece calculates a score for each character pair, prioritizing pairs more likely to appear together. This avoids a rigid focus on the most frequently occurring pairs, fostering improved tokenization. Despite shared characteristics with BPE, WordPiece stands out for its resource efficiency, making it suitable for algorithms demanding fewer computational resources, as it does not store the merges. 
    \item \textbf{Unigram:} Contrary to its counterparts, Unigram aims to identify the most common sub-strings in the corpus using frequency analysis or even applying BPE initially. Unigram Tokenization then seeks optimal tokenization for each word, calculating probabilities based on sub-string probabilities. By iteratively removing tokens with minimal impact on overall loss, the algorithm refines the vocabulary. Despite yielding more logical sub-strings \cite{Bostrom2020BytePE}, Unigram Tokenization demands significantly more time and computational resources compared to BPE or WordPiece.
\end{itemize}

\subsection{Evaluation Methods}
In order to evaluate the embedding and tokenization methods, we do need to train downstream models in order to either calculate similarity, predict the entity, or perform classification. In each of the tasks, the dataset is first embedded using the relevant embedding model.

\subsubsection{Word and Sentence Similarity}
\label{subsec:ws_model}
We calculate cosine similarity using the embedding model that we are currently using against the ground truth. In case of an out-of-vocabulary word, depending on the tokenization method, the embedding would be set to zeros (word tokenization) or it would be set to the mean of the sub-words/characters of that word (sub-word tokenization). For metrics, we calculate Pearson, Kendall, and Spearman correlation between the cosine similarity from the embedding model and the human-annotated similarity. 

In sentence similarity, a sentence representation is carried out as the mean of all the token representations contained within that sentence. After that, the same process as in the case of word similarity is carried out to evaluate the performance of sentences along with the same metrics as described above.

\subsubsection{Named Entity Recognition}
\label{subsec:ner_model}
For evaluating the embeddings for NER, we create a simple two layer bi-directional LSTM \cite{Hochreiter1997LongSM} with one fully connected layer that outputs the probability of each token using a softmax activation \cite{LeCun2015DeepL}. The Neural Networks is trained for 5 epochs with a learning rate of 0.0005 and batch size of 128. The evaluation is carried out using various classification metrics, which include accuracy, precision, recall, f-beta, and matthews correlation coefficient \cite{Matthews1975ComparisonOT}.

\subsubsection{Document Classification}
\label{subsec:dc_model}
For document classification, we start by simply creating a dataset schema that stores the start and end index of the text along with the label class. We then use our embedding model which embeds the whole text and calculates a mean embedding of the text. It then uses a KNN-classifier \cite{Cover1967NearestNP} with neighbors, \textit{k} = 3, to predict the output class for a given text. The evaluation is done using the same classification metrics as NER.

\subsection{Training Datasets}
\label{subsec: train_ds}
We now describe the datasets that these models were trained on. We choose the datasets for our purposes: 1) for general purpose embedding generation and 2) and scientific domain specific embedding generation.
\subsubsection{General Domain Datasets}
\label{subsubsec: wiki_ds}
For generic word embeddings, we use pre-trained embeddings of Word2Vec \footnote{\url{https://nlp.stanford.edu/projects/glove/}} and FastText \footnote{\url{https://fasttext.cc/docs/en/english-vectors.html}} available on the relevant websites and already trained on huge corpora of general-domain data

\subsubsection{Scientific Domain Dataset}
\label{subsubsec: abstracts_ds}
For scientific domain specific embeddings' generation, we are using an Iris.AI \footnote{\url{https://iris.ai/}} company confidential dataset. The dataset contains around ~2.5 million abstracts, with a mean length of around ~73 words. These abstracts are selected from 230 million open access articles hosted by the CORE group \footnote{\url{https://core.ac.uk/}}. The dataset contains English abstracts of scientific articles from different fields of research. Each of the abstracts is lower-cased along with stopwords and punctuation removed. During tokenization, bigrams are constructed. In addition, scientific entities, such as chemical formulas and numbers with units, are masked to their normalized forms to increase their representations. Each token has also been tagged with their part-of-speech (POS). For clarification, we will refer to this dataset as \textbf{Abstracts} dataset from now on.

\section{Results}
The results for embedding evaluation are summarized in Table \ref{tab:best_results}. For detailed results, refer to Appendix \ref{sec:results_appendix}

\begin{table*}[t]
\centering
\begin{tabular}{@{}lllll@{}}
\toprule
\textbf{Task}               & \textbf{Dataset}  & \textbf{Metric} & \textbf{Best Model}          & \textbf{Best Result }\\ \midrule
Word Similarity             & UNMSRS            & Pearson         & W2V Abstracts 100D Skipgram  & 0.5032 \\
Sentence Similarity         & SemEval           & Pearson         & W2V Abstracts 200D Skipgram  & 0.6749 \\
Sentence Similarity         & Clinical STS 2018 & Pearson         & W2V Abstracts 200D Skipgram  & 0.7588 \\
Sentence Similarity         & Clinical STS 2019 & Pearson         & W2V Abstracts 200D Skipgram  & 0.6362 \\
Named Entity Recognition    & Conll 2003        & F-Beta          & RoBERTa                      & 0.8307 \\
Named Entity Recognition    & CHEMDNER          & F-Beta          & SciBERT                      & 0.5399 \\
Named Entity Recognition    & SciERC            & F-Beta          & SciBERT                      & 0.3841 \\
Document Classification     & Reuters 12        & F-Beta          & W2V Abstracts 200D Skipgram  & 0.9243 \\
Document Classification     & BioChem 8         & F-Beta          & W2V Abstracts 200D Skipgram  & 0.9015 \\
\bottomrule
\end{tabular}
\caption{Best Results on each individual task and dataset, for embedding evaluation without tokenization}
\label{tab:best_results}
\end{table*}

\subsection{Embedding Evaluation}

\subsubsection{Word Similarity}
The results of the word representation models on the word similarity dataset, UNMSRS, show that Word2Vec models based on skipgram and trained on the scientific domain dataset performs the best, but that is for the in-vocabulary tokens i.e, the OOV tokens are completely ignored when calculating the result. In our case, 189 pairs out of the 566 had one or both of the tokens as OOV, for which the model produced NaN values (More on it in section \ref{subsec: tok_eval}). Notably, FastText models trained on the scientific domain dataset perform similarly well, offering generalization for OOV tokens. In contrast, transformer-based models, including SciBERT, perform poorly as they rely on context, which is lacking in word similarity tasks. Refer to Table \ref{tab:unmsrs_results} in Appendix \ref{sec:results_appendix} for detailed results from all models.

\subsubsection{Sentence Similarity}
The Word2Vec model trained on the Scientific Domain dataset using the Skipgram method demonstrates superior performance across various sentence similarity datasets.

For all three datasets, the Word2Vec model trained on the abstracts dataset with skipgram and a 200-dimensional vector excels, closely followed by its 100-dimensional counterpart (refer to Table \ref{tab:semeval_results} in Appendix \ref{sec:results_appendix}). The model's performance on the scientific specific datasets (STS 2018 and 2019) was expected due to the same domain but the model performs impressively even on a generic domain dataset (SemEval), possibly influenced by the diverse data sources like newspapers, open forums, and tweets within the SemEval dataset, which could also include scientific data. Transformer-based models, benefiting from contextual information, exhibit slight improvements, with general-purpose models like BERT and RoBERTa outshining SciBERT in case of SemEval and SciBERT out-performing in case of scientific datasets. Interestingly, FastText's comparatively weaker performance may be attributed to its reliance on n-gram knowledge, which might not always be suitable, especially with less logical information for certain words.


These findings highlight the robustness of the Word2Vec model trained on the Scientific Domain dataset across different sentence similarity datasets, emphasizing its efficacy in capturing domain-specific nuances.

\subsubsection{Named Entity Recognition}
In the extrinsic evaluation task, Named Entity Recognition (NER), transformer-based models exhibit superior performance compared to other word representation models. This is logical given the extensive contextual information available in longer documents, together with the utilization of Neural Networks (LSTMs) as the model used for predicting the entities.

RoBERTa emerges as the top performer among word representation models in generic NER on the CONLL 2003 dataset, closely followed by BERT and GPT-2, as detailed in Table \ref{tab:conll_results} in Appendix \ref{sec:results_appendix}. Again, this is in-line with our initial hypothesis, as these models are pre-trained on generic data, indicating their effectiveness on general-purpose datasets. Interestingly, SciBERT outperforms Word2Vec and FastText models trained on Wikipedia data, once again underscoring the generalization capabilities of transformer-based models. 

In the domain of scientific NER, SciBERT consistently outperforms other models, as evidenced by the results on ChemDNER and SciERC (Tables \ref{tab:chemdner_results} and \ref{tab:scierc_results} respectively in Appendix \ref{sec:results_appendix}). Also, SciBERT's superior performance on the chemistry-specific ChemDNER dataset is intriguing, considering its pre-training on a more general scientific corpus. Even BERT and RoBERTa exhibit better performance than Word2Vec and FastText models trained on scientific-specific data, emphasizing the strength of context-based word embeddings generated by transformer-based models.

These findings underscore the effectiveness of transformer-based word representation models in capturing nuanced information for Named Entity Recognition tasks, both in generic and scientific domains.

\subsubsection{Document Classification}
One of the most interesting outcome were the results related to document classification, with the Word2Vec model trained on scientific domain data emerging as the top performer, not only on scientific-specific datasets but also on general-purpose data, demonstrating the models' generalization capabilities.

Word2Vec model training \textit{abstracts} dataset using the Skipgram method with a 200-dimensional dataset performed the best on both the Reuters 12, a generic domain document classification dataset, and BioChem 8, a scientific-specific document classification dataset (results in Tables \ref{tab:reuters12_results} and \ref{tab:biochem_results} respectively in Appendix \ref{sec:results_appendix}). These results affirm the models' proficiency in both general and scientific document classification scenarios.

\subsection{Tokenization Evaluation}
\label{subsec: tok_eval}
We can see the benefits of tokenization in terms of transformer-based models performing well on some tasks, as compared to Word2Vec models. FastText also encodes the sub-word information so it is able to handle OOV words as well. What we have not analyzed is the benefits that tokenization allows in particular, irrespective of the word representation model used. One of the reason that Word2Vec model performs so good here is due to no limit on the vocabulary size i.e, currently when Word2Vec model is trained on \textit{abstracts} dataset, it has a vocabulary size of 1.4 million which is almost 50 times more than standard vocabulary size of 50,000 for sub-word tokenizers. In such a case, where there is no limit of vocabulary size, word-based tokenizers are expected to perform better than sub-word tokenizers. In this section, we will solely focus on how sub-word tokenization can help instead of using word-based tokenizers. We will focus on all the tasks and see which one benefit from sub-word tokenization. Only in the case of Word Similarity, we are strictly looking at the performance of the embedding model on OOV words tokens and generate similarity metrics for those specific tokens. In other tasks, we look at overall metrics of the specific task. Detailed results are given in the appendix section \ref{sec:tok_results}. \\

For the evaluation purposes, we first train a subword tokenization model on the \textit{abstracts} dataset, for it to learn the tokenization. In total, we run the following experiments, all of them are run both with cbow and skipgram, and have a dimension of 100:

\begin{itemize}
   \item Word2Vec model trained on individual words.
    \item Word2Vec model trained on sub-words with three different tokenization methods i.e, BPE, Unigram, and WordPiece.
    \item FastText model trained on \textit{Abstracts} dataset.
\end{itemize}

\subsubsection{Word Similarity}
\label{subsubsec:tok_ws}

\begin{table*}[t]
\centering
\begin{tabular}{@{}llllll@{}}
\toprule
 &
   &
   &
  \multicolumn{1}{c}{\textbf{In Vocab}} &
  \multicolumn{1}{c}{\textbf{OOV}} &
  \multicolumn{1}{c}{\textbf{Overall}} \\ \midrule
\textbf{Embedding   Model} &
  \textbf{Algorithm} &
  \textbf{Tokenization Method} &
  \textbf{Pearson} &
  \textbf{Pearson} &
  \textbf{Pearson} \\ \midrule
\textbf{Word2Vec} & Cbow     & -         & 0.4848          & NaN             & -      \\
\textbf{Word2Vec} & Skipgram & -         & 0.5115          & NaN             & -      \\
\textbf{}         &          &           &                 & \textbf{}       &        \\
\textbf{Word2Vec} & Cbow     & BPE       & 0.2521          & 0.2199          & 0.2419 \\
\textbf{Word2Vec} & Cbow     & Unigram   & 0.0878          & 0.2232          & 0.1362 \\
\textbf{Word2Vec} & Cbow     & WordPiece & 0.3044          & 0.2862          & 0.2982 \\
\textbf{}         &          &           &                 &                 &        \\
\textbf{Word2Vec} & Skipgram & BPE       & 0.2940          & 0.2348          & 0.2751 \\
\textbf{Word2Vec} & Skipgram & Unigram   & 0.2193          & 0.2484          & 0.2275 \\
\textbf{Word2Vec} & Skipgram & WordPiece & 0.3660          & 0.2897          & 0.3412 \\
                  &          &           &                 &                 &        \\
\textbf{FastText} & Cbow     & N-grams   & 0.5318          & \textbf{0.3662} & 0.4683 \\
\textbf{FastText} & Skipgram & N-grams   & \textbf{0.5405} & 0.3654          & \textbf{0.4737} \\ \bottomrule
\end{tabular}
\caption{Results of Tokenization Evaluation on UNMSRS Word Similarity Dataset. The dataset contains 189 OOV out of 566 Pairs}
\label{tab:tok_ws}
\end{table*}

We will start with Word Similarity. In this, we can accurately measure how sub-word tokenizers deal with OOV tokens. We compare different tokenization methods with Word2Vec along with FastText Model. Currently, our UNMSRS dataset contains 189 OOV out of 566 pairs, i.e, at least one of the words those pairs is OOV. \\

Table \ref{tab:tok_ws} shows the results of this evaluation. As expected, for the word based tokenizer, the score is \textit{NaN} for the OOV tokens, as it is unable to generate an embedding for it. It is able to generate a relatively good pearson score for in vocabulary tokens though. On the other hand, sub-word tokenizer, with Word2Vec, tends to generate average scores for these tokens but gives a better score than NaN for OOV tokens, albeit not a really good one. Finally, FastText tends to perform the best in this case, as it gives the best scores for both in-vocabulary and OOV tokens. As the correlation metrics ignore NaN values, the Table \ref{tab:best_results} seems to show that Word2Vec out-performs other methods on the UNMSRS dataset, but that is because it ignores NaN values, whereas FastText combines results of both. \\

\subsubsection{Sentence Similarity}
\label{subsubsec:tok_ss}
Table \ref{tab:tok_ss} in Appendix \ref{sec:results_appendix} shows that tokenization particularly does not give a rise to the scores on sentence similarity datasets, where simple Word2Vec models based on word-tokenizers tend to out-perform all subword tokenizers. This could be due to the reason that Word2Vec model based on word-tokenizers already have a vocabulary size of 1.4M, so there is a less chance of a word being OOV in a given sentence and even if there is, the overall context of most sentences can be derived even if some of them are OOV.

\subsubsection{Document Classification}
\label{subsubsec:tok_dc}
Similar to the results of Sentence Similarity, the word-based tokenizers out-perform subword tokenization in the case of document classification as well, as seen in Table \ref{tab:tok_dc}. Again, the reason could be similar that the context of a document (the output class) can be inferred from the whole text, even if we haven't seen some of the words before. 

\subsubsection{Named Entity Recognition}
\label{subsubsec:tok_ner}
Here, we see the improvement of tokenization as the models based on sub-word tokenization, especially WordPiece tend to out-perform all other methods, on all the NER datasets. Also, the Word2Vec models based on all 3 tokenization methods out-perform Word2Vec based on word-based models, atleast when the underlying algorithm is Cbow. What is interesting to see is that in case of SciERC dataset, the WordPiece tokenization method seems to approach the result achieved by SciBERT (0.3254 vs 0.3841) (see Table \ref{tab:scierc_results} in Appendix \ref{sec:results_appendix}), showing the power of tokenization. The reason why tokenization works on NER and not on Sentence Similarity and Document Classification, is because we are actually focusing on individual words when predicting the entity, where tokenization is helpful, especially if the entity to be predicted is OOV, which would never be predicted by a word-representation model without tokenization

\section{Conclusion}
\label{subsec:conclusion}

In this paper, we investigated methods for word representation and tokenization, particularly within the scientific domain, aiming to determine their effects on downstream NLP tasks. We compared state-of-the-art Transformer models with simpler models like Word2Vec and FastText, considering computational efficiency and performance on domain-specific data. We developed an evaluation suite encompassing diverse tasks and datasets, including intrinsic and extrinsic evaluations. Word2Vec performed strongly across most tasks, particularly in intrinsic evaluation, sometimes outperforming Transformer models in extrinsic evaluation, emphasizing the effectiveness of whole-word representations over sub-word tokenization, especially for in-vocabulary words, and where there is a lack of context (similarity tasks). We conclude that a hybrid approach combining word-based tokenization with Word2Vec alongside sub-word tokenization, specifically to handle out-of-vocabulary tokens, can be used achieve comparable performance to Transformer models with significantly fewer resources and parameters. One area of further research could be fine-tuning the word2vec embeddings on downstream tasks as well.

\section*{Limitations}
This study, however, has some limitations. Firstly, the results go against the current state-of-the-art (Generative AI and Transformer Based Models), and secondly, require a lot more experimentation to be considered really reliable and generalizable. For instance, experiments are required for testing with various downstream tasks, along with different downstream models for prediction to see if there's a difference between using certain word-representation model for a specific downstream task/model. Also, for a fair comparison, there needs to pre-training/fine-tuning on domain specific data for all the word-representation models, in this case Transformer-Based models. Of course, that would result in a lot of computational power, which is not readily available but would be necessary for these findings to stand. In addition to this, a detailed use of computational power and time for each algorithm could help in deciding whether a certain improvement in results is worth the resources used. \\ 
\bibliography{anthology,custom}
\bibliographystyle{acl_natbib}

\onecolumn
\appendix

\section{Appendix}
\label{sec:eval_framework_appendix}
Below is in detail description of the tasks and their relevant datasets for our evaluation framework:
\subsection{Word Similarity}
\label{subsec:wordsim_task}
Word similarity is one of the basic building blocks of a lot of Natural language processing (NLP) tasks as it captures how and if the word is correctly represented in general and specifically within a particular context as well. Word similarity can be used to gauge how well the embeddings are able to extract the semantics of the words given the current context. The reason behind choosing word similarity as a task in our evaluation suite is two fold: 1) It is one of the building blocks for any kind of intrinsic evaluation and 2) it helps us to get an idea on how the model will further perform on more useful downstream tasks such as extractive summarization, which basically works by extracting the semantics of sentences and clustering them based on their significance in a document. 

\subsubsection{UMNSRS}
\label{subsubsec:umnsrs_ws}
University of Minnesota Semantic Relatedness Standard (UMNSRS) (\cite{Pakhomov2018SemanticRA}) is a collection of concept pairs that are manually annotated for both similarity and relatedness using a continuous response scale. Here, we mainly focus on conducting the word similarity. The term pairs have been evaluated by various healthcare professionals (e.g., medical coders, residents, clinicians) to determine the degree of semantic similarity. This dataset is useful in our evaluation suite as it gives us a basic understanding of the word embeddings on a token level, specific to the scientific domain, which serves as a building block for our evaluation suite. The dataset contains a total of 566 examples and has a mean length of 9.12 for each word.

\subsection{Sentence Similarity}
\label{subsec:sentencesim_task}
Sentence Similarity builds on the task of word similarity and determines how similar two texts are. It works on basically capturing the semantic information of the texts and is particularly useful for information retrieval tasks and clustering/grouping. The reason for choosing this task in our evaluation suite is similar to Word Similarity with the addition that it could be also useful in downstream tasks asuch as summarization where it can be used to either remove redundant information or extracting useful information which is re-iterated multiple times in the given text. The score in our datasets range from 0 to 5, with the explanation of each pair of sentences for all the datasets below, and they are annotated using the following explanations:
\begin{itemize}
    \item \textit{5: } The two sentences are completely equivalent, as they mean the same thing.
    \item \textit{4: }The two sentences are mostly equivalent, but some unimportant details differ.
    \item \textit{3: }The two sentences are roughly equivalent, but some important information differs/missing.
    \item \textit{2: }The two sentences are not equivalent, but share some details.
    \item \textit{1: }The two sentences are not equivalent, but are on the same topic.
    \item \textit{0: }The two sentences are completely dissimilar.
\end{itemize}

\subsubsection{SemEval-2017}
\label{subsubsec:semeval_ss}
SemEval is an international workshop and a shared task (\cite{cer-etal-2017-semeval}) on Semantic Evaluation held every year which aims to measure semantic textual similarity (STS) given a dataset. The dataset comprises a selection of the English datasets used in the STS tasks organized in the context of SemEval between 2012 and 2017. The selection of datasets include text from image captions, news headlines and user forums. Evaluation on this dataset gives us an understanding of our word representation model on how it performs on general purpose data. The dataset contains a total of 8628 examples with the mean length of each sentence being 10.17 words.

\subsubsection{Clinical STS 2018 | 2019}
\label{subsubsec:sts_ss}
Building on this, we require a scientific specific dataset as well, to compare the performance of our tokenization and word representation models. Introduced as a shared task in SemEval, in an attempt to improve the systems for semantic textual similarity specific to medical data (\cite{Wang2018MedSTSAR}). The organizers, collected data from Mayo Clinic’s clinical data warehouse (\cite{Horton2017EmpoweringMC}), applied sentence pairs selection among different categories of unified medical language system (UMLS) \footnote{https://www.nlm.nih.gov/research/umls/} semantic types: sign and symptom, disorder, procedure and medication. Two annotators, having vast experience of many years in clinical domain, annotated the sentences manually, with weighted Cohen's Kappa of 0.67. The datasets contains a total of 1068 and 2054 examples with the mean lengths of each sentence being 25.42 and 19.35 words for the 2018 and 2019 versions of the dataset respectively.

\subsection{(Named) Entity Recognition}
\label{subsec:ner_task}
Named Entity Recognition (NER) (\cite{Mohit2014NamedER}) is the process of extracting entities (particularly named entities) from unstructured text and then classifying them into different pre-defined categories such as person names, organizations, locations, and so on. This is the starting point of our extrinsic evaluation and the reason behind is identifying the different entities that can be found in our text, specifically in terms of Scientific reports. For instance, it is particularly important to extract quantities of certain chemicals from a medical report, and extracting the chemicals (entities) will help us in doing that. Below is a simple annotated example:

\begin{quote}
    Jim bought 300 shares of Acme Corp. in 2006.
\end{quote}

\begin{quote}
    $[Jim]_{Person}$ bought 300 shares of $[Acme Corp.]_{Organization}$ in $[2006]_{Time}$.
\end{quote}


\subsubsection{CoNLL-2003}
\label{subsubsec:conll_ner}
CoNLL-2003 dataset (\cite{Sang2003IntroductionTT}) was introduced as a shared task. The dataset in itself contains three separate tasks, which includes part-of-speech tagging, chunking (which refers to grouping similar parts of text), and named entity recognition (NER). Here, we mainly focus on NER, which concentrates on four types of named entities: persons, locations, organizations, and names of miscellaneous entities that do not belong to the previous three groups. The named entity tags have the format I-TYPE which means that the word is inside a phrase of type TYPE. Only if two phrases of the same type immediately follow each other, the first word of the second phrase will have tag B-TYPE to show that it starts a new phrase. A word with tag O is not part of a phrase. This dataset serves as a starting point for extrinsic evaluation in our suite. As this is a general purpose dataset, we can compare different embedding models trained on scientific as well as generic data.

\subsubsection{ChemdNER}
\label{subsubsec:chemdner_ner}
CHEMDNER (\cite{Krallinger2015TheCC}) is chosen as another NER dataset for our evaluation suite as it covers the scientific domain, specifically chemistry. Moreover, the entity classes in this dataset are more general, like Abbreviations, which are common in any scientific data. CHEMDNER is a collection 10,000 PubMED abstracts, containing a total of 84,355 chemical entity mentions labeled manually by expert chemistry literature  curators. The chemical entities were labeled according to their structure-associated chemical entity mention (SACEM) class: 

\begin{itemize}
    \item \textit{Abbreviation: } Abbreviations of acronyms of chemicals/drugs.
    \item \textit{Identifiers: } Database identifiers, PubChem ids.
    \item \textit{Formula: } Molecular formulas
    \item \textit{Systematic: } Systematic names: IUPAC or IUPAC-like (The International Union of Pure and Applied Chemistry).
    \item \textit{Multiple: } Chemical mentions that are not described in a continuous string of characters.
    \item \textit{Trivial: } Trivial, trade, brand, common or generic names.
    \item \textit{Family: } Chemical families that can be linked to structures.
\end{itemize}

\subsubsection{SciERC}
\label{subsubsec:scierc_ner}
Building on this, we choose another scientific domain dataset for NER but this time more focused to the computer science, specifically Artificial Intelligence (AI) domain. SciERC (\cite{Luan2018MultiTaskIO}) was introduced as a multi-task setup in order to identify entities, relations, and co-reference clusters in scientific articles. Here, we mainly focus on identifying entities. It includes scientific entities for 500 scientific abstracts that are taken from 12 AI conference/workshop proceedings in four AI communities, from the Semantic Scholar Corpus \cite{Lo2020S2ORCTS}. The different entity categories are mentioned below:

\begin{itemize}
    \item \textit{Task: }Applications, problems to solve, systems to construct. E.g. information extraction, machine reading system, image segmentation, etc.
    \item \textit{Method: }Methods , models, systems to use, or tools, components of a system, frameworks. E.g. language model, CORENLP, POS parser, kernel method, etc.
    \item \textit{Evalution Metric: }Metrics, measures, or entities that can express quality of a system/method. E.g. F1, BLEU, Precision, Recall, ROC curve, mean reciprocal rank, mean-squared error, robustness, time complexity, etc.
    \item \textit{Material: }Data, datasets, resources, Corpus, Knowledge base. E.g. image data, speech data, stereo images, bilingual dictionary, paraphrased questions, CoNLL, Panntreebank, WordNet, Wikipedia, etc.
    \item \textit{Other Scientifc Terms: }Phrases that are a scientific terms but do not fall into any of the above classes. E.g. physical or geometric constraints, qualitative prior knowledge, discourse structure, syntactic rule, discourse structure, tree, node, tree kernel, features, noise, criteria, etc.
    \item \textit{Generic: }General terms or pronouns that may refer to a entity but are not themselves informative, often used as connection words. E.g model, approach, prior knowledge, them, it...
\end{itemize}

Note that the entity classes in the two scientific specific datasets are quite different. This is particularly useful for us as we have a more robust set of evaluation datasets in our suite. Table \ref{tab:ner_stats} shows the stats for the three NER datasets described above.

\begin{table}[h!]
\centering

\begin{tabular}{@{}llll@{}}
\toprule
\textbf{Dataset Name} & \textbf{Number of Sentences} & \textbf{Number of Entities} & \textbf{Entities Per Class}                                                                       \\ \midrule
Conll 2003   & 13151               & 35089              & \begin{tabular}[c]{@{}l@{}}LOC: 10645\\ PER: 10059\\ ORG: 9323\\ MISC: 5062\end{tabular} \\ \midrule
ChemdNER &
  31465 &
  29478 &
  \begin{tabular}[c]{@{}l@{}}Trivial: 8832\\ Systematic: 6656\\ Abbreviation: 4538\\ Formula:   4448\\ Family: 4090\\ Identifier: 672\\ Multiple: 202\\ No Class: 40\end{tabular} \\ \midrule
SciErc &
  2717 &
  8098 &
  \begin{tabular}[c]{@{}l@{}}OtherScientificTerm: 2271\\ Method: 2094\\ Generic: 1338\\ Task:   1284\\ Material: 771\\ Metric: 340\end{tabular} \\ \bottomrule
\end{tabular}
\caption{Dataset Statistics for Named Entity Recognition Datasets}
\label{tab:ner_stats}
\end{table}
\subsection{Document Classification}
\label{subsec:dc_task}
Document classification (\cite{Wan2019LonglengthLD}) refers to the process of assigning a document to relevant pre-defined categories for easier further management of analysis. There are several reasons behind choosing this as part of our extrinsic evaluation. For instance, in information retrieval systems, such as search engines, automatic document classification (particularly using machine learning) techniques are extremely important and useful, making it easier and a smooth process for the users, to find what they are looking for. It also makes it easier to filter and analyze documents using the assigned categories. The use of document classification in such systems makes it important for us to consider it for our evaluation suite.

\subsubsection{Reuter 12}
\label{subsubsec:reuters_dc}
Reuters \footnote{\url{https://martin-thoma.com/nlp-reuters/}} is a benchmark dataset for document classification. Originally, it is a multi-class and multi-label (each document can have more than one class) dataset. It has 90 classes, 7769 training documents, and 3019 testing documents. For our purposes, we selected a subset of the dataset. The subset of the dataset contained 12 classes and is a single label, multi-class dataset meaning that there are multiple classes but each document belongs to only one class. The final dataset contained 3368 documents and the final classes, with their counts can be seen in Table \ref{tab:dc_stats}.

\subsubsection{Biochem 8}
\label{subsubsec:corebiochem_dc}
Biochem dataset \footnote{\url{https://github.com/xdwang0726/MeSHup}} is a document classification dataset containing biomedical documents classified into 8 different classes. The dataset is a multi-class, single label dataset. The final dataset contains 39845 documents and the final classes, with their counts can be seen in Table \ref{tab:dc_stats}.

\begin{table}[h!]
\centering

\begin{tabular}{@{}llll@{}}
\toprule
\textbf{Dataset Name} &
  \textbf{Number of Documents} &
  \textbf{Number of Classes} &
  \textbf{Examples Per Class} \\ \midrule
Reuters News &
  3368 &
  12 &
  \begin{tabular}[c]{@{}l@{}}acq: 1809\\ crude: 349\\ trade: 330\\ interest: 164\\ ship: 151\\ sugar: 127\\ coffee: 107\\ gold: 97\\ gnp: 72\\ cpi: 65\\ cocoa: 52\\ iron-steel: 45\end{tabular} \\ \midrule
Bio Chem &
  39845 &
  8 &
  \begin{tabular}[c]{@{}l@{}}anatomy: 5000\\ abortion: 5000\\ cancer: 5000\\ microbiology: 5000\\ disease:   4983\\ epigenetics: 4978\\ biochemistry: 4956\\ immunology: 4928\end{tabular} \\  \bottomrule
\end{tabular}
\caption{Dataset Statistics for Document Classification Datasets}
\label{tab:dc_stats}
\end{table}

\clearpage

\section{Appendix}
\label{sec:results_appendix}
\subsection{Results}

Below are the detailed results for each task and their corresponding datasets:

\subsubsection{Word Similarity}
\begin{table}[t]
\centering
\begin{tabular}{@{}llll@{}}
\toprule
\textbf{Model Name}                          & \textbf{Pearson} & \textbf{Kendall} & \textbf{Spearman} \\ \midrule
W2V Wiki & 0.3573 & 0.2240 & 0.3290 \\
W2V Abstracts 100D Cbow & 0.4146 & 0.2729 & 0.3981 \\
W2V Abstracts 200D Cbow & 0.4320 & 0.2802 & 0.4083 \\
\textbf{W2V Abstracts 100D Skipgram} & \textbf{0.5032} & \textbf{0.3391} & \textbf{0.4874} \\
W2V Abstracts 200D Skipgram & 0.5027 & 0.3350 & 0.4822 \\
FT Wiki & 0.3947 & 0.2603 & 0.3834 \\
FT Abstracts 100D Cbow & 0.4772 & 0.3128 & 0.4531 \\
FT Abstracts 200D Cbow & 0.4787 & 0.3118 & 0.4484 \\
FT Abstracts 100D Skipgram & 0.4813 & 0.3164 & 0.4575 \\
FT Abstracts 200D Skipgram & 0.4824 & 0.3159 & 0.4532 \\
BERT & 0.1662 & 0.0995 & 0.1491 \\
RoBERTa & 0.1888 & 0.1132 & 0.1683 \\
GPT-2 & 0.0491 & 0.0489 & 0.0738 \\
SciBERT & 0.2048 & 0.1199 & 0.1766 \\ \bottomrule
\end{tabular}
\caption{Results of Word Similarity on UNMSRS Dataset}
\label{tab:unmsrs_results}
\end{table}
\newpage

\subsubsection{Sentence Similarity}
\begin{table}[h!]
\centering
\begin{tabular}{@{}llll@{}}
\toprule
\textbf{Model Name}                          & \textbf{Pearson} & \textbf{Kendall} & \textbf{Spearman} \\ \midrule
W2V Wiki & 0.4270 & 0.3341 & 0.4763 \\
W2V Abstracts 100D Cbow & 0.5967 & 0.4135 & 0.5739 \\
W2V Abstracts 200D Cbow & 0.6098 & 0.4225 & 0.5848 \\
W2V Abstracts 100D Skipgram & 0.6567 & 0.4771 & 0.6499 \\
\textbf{W2V Abstracts 200D Skipgram} & \textbf{0.6749} & \textbf{0.4859} & \textbf{0.6601} \\
FT Wiki & 0.5485 & 0.4371 & 0.6068 \\
FT Abstracts 100D Cbow & 0.4300 & 0.3218 & 0.4560 \\
FT Abstracts 200D Cbow & 0.4511 & 0.3301 & 0.4672 \\
FT Abstracts 100D Skipgram & 0.4310 & 0.3215 & 0.4556 \\
FT Abstracts 200D Skipgram & 0.4490 & 0.3292 & 0.4659 \\
BERT & 0.5262 & 0.3567 & 0.5078 \\
RoBERTa & 0.5422 & 0.3953 & 0.5549 \\
GPT-2 & 0.2033 & 0.2153 & 0.3128 \\
SciBERT & 0.4705 & 0.3346 & 0.4784 \\ \bottomrule
\end{tabular}
\caption{Results of Sentence Similarity on SemEval Dataset}
\label{tab:semeval_results}
\end{table}
\begin{table}[h!]
\centering
\begin{tabular}{@{}llll@{}}
\toprule
\textbf{Model Name}                          & \textbf{Pearson} & \textbf{Kendall} & \textbf{Spearman} \\ \midrule
W2V Wiki & 0.6001 & 0.4141 & 0.5738 \\
W2V Abstracts 100D Cbow & 0.7349 & 0.4693 & 0.6352 \\
W2V Abstracts 200D Cbow & 0.7450 & 0.4723 & 0.6385 \\
W2V Abstracts 100D Skipgram & 0.7514 & 0.4715 & 0.6356 \\
\textbf{W2V Abstracts 200D Skipgram} & \textbf{0.7588} & \textbf{0.4716} & \textbf{0.6353} \\
FT Wiki & 0.4686 & 0.3490 & 0.4780 \\
FT Abstracts 100D Cbow & 0.4382 & 0.3387 & 0.4732 \\
FT Abstracts 200D Cbow & 0.4449 & 0.3412 & 0.4773 \\
FT Abstracts 100D Skipgram & 0.4363 & 0.3389 & 0.4735 \\
FT Abstracts 200D Skipgram & 0.4421 & 0.3423 & 0.4784 \\
BERT & 0.6613 & 0.3884 & 0.5317 \\
RoBERTa & 0.6143 & 0.4219 & 0.5736 \\
GPT-2 & 0.1557 & 0.2812 & 0.3908 \\
SciBERT & 0.6752 & 0.4292 & 0.5858 \\ \bottomrule
\end{tabular}
\caption{Results of Sentence Similarity on Clinical STS 2018 Dataset}
\label{tab:sts_2018_results}
\end{table}
\begin{table}[h!]
\centering
\begin{tabular}{@{}llll@{}}
\toprule
\textbf{Model Name}                          & \textbf{Pearson} & \textbf{Kendall} & \textbf{Spearman} \\ \midrule
W2V Wiki & 0.5852 & 0.4538 & 0.6274 \\
W2V Abstracts 100D Cbow & 0.5975 & 0.4360 & 0.5880 \\
W2V Abstracts 200D Cbow & 0.6087 & 0.4408 & 0.5936 \\
W2V Abstracts 100D Skipgram & 0.6300 & 0.4629 & 0.6221 \\
\textbf{W2V Abstracts 200D Skipgram} & \textbf{0.6362} & \textbf{0.4631} & \textbf{0.6217} \\
FT Wiki & 0.4881 & 0.4043 & 0.5686 \\
FT Abstracts 100D Cbow & 0.4031 & 0.3300 & 0.4554 \\
FT Abstracts 200D Cbow & 0.4209 & 0.3376 & 0.4661 \\
FT Abstracts 100D Skipgram & 0.4056 & 0.3311 & 0.4571 \\
FT Abstracts 200D Skipgram & 0.4196 & 0.3375 & 0.4656 \\
BERT & 0.5307 & 0.3821 & 0.5171 \\
RoBERTa & 0.5423 & 0.4046 & 0.5460 \\
GPT-2 & 0.2214 & 0.2903 & 0.4020 \\
SciBERT & 0.5593 & 0.3961 & 0.5382 \\ \bottomrule
\end{tabular}
\caption{Results of Sentence Similarity on Clinical STS 2019 Dataset}
\label{tab:sts_2019_results}
\end{table}
\newpage

\subsubsection{Named Entity Recognition}
\begin{table}[h!]
\centering

\begin{tabular}{@{}llllll@{}}
\toprule
\textbf{Model Name} & \textbf{Accuracy} & \textbf{Precision} & \textbf{Recall} & \textbf{F\_beta} & \textbf{Matthews} \\ \midrule
W2V Wiki & 0.7737 & 0.4901 & 0.7879 & 0.5852 & 0.5475 \\
W2V Abstracts 100D Cbow & 0.6450 & 0.3041 & 0.6513 & 0.3973 & 0.3893 \\
W2V Abstracts 200D Cbow & 0.6589 & 0.3245 & 0.6752 & 0.4180 & 0.4053 \\
W2V Abstracts 100D Skipgram & 0.6628 & 0.3107 & 0.6351 & 0.4040 & 0.3979 \\
W2V Abstracts 200D Skipgram & 0.6675 & 0.3454 & 0.6741 & 0.4350 & 0.4137 \\
FT Wiki & 0.1011 & 0.1623 & 0.2912 & 0.1451 & 0.0828 \\
FT Abstracts 100D Cbow & 0.6510 & 0.3188 & 0.6508 & 0.4097 & 0.4022 \\
FT Abstracts 200D Cbow & 0.7232 & 0.3706 & 0.6972 & 0.4684 & 0.4700 \\
FT Abstracts 100D Skipgram & 0.6764 & 0.3373 & 0.6479 & 0.4238 & 0.4186 \\
FT Abstracts 200D Skipgram & 0.7131 & 0.3718 & 0.6956 & 0.4663 & 0.4615 \\
BERT & 0.8952 & 0.7230 & 0.8901 & 0.7853 & 0.7709 \\
\textbf{RoBERTa} & \textbf{0.9111} & \textbf{0.7795} & \textbf{0.9155} & \textbf{0.8307} & \textbf{0.8155} \\
GPT-2 & 0.8905 & 0.7482 & 0.8220 & 0.7673 & 0.7691 \\
SciBERT & 0.8672 & 0.6904 & 0.8508 & 0.7463 & 0.7354 \\ \bottomrule
\end{tabular}
\caption{Results of Named Entity Recognition on Conll 2003 Dataset}
\label{tab:conll_results}
\end{table}
\begin{table}[h!]
\centering

\begin{tabular}{@{}llllll@{}}
\toprule
\textbf{Model Name} & \textbf{Accuracy} & \textbf{Precision} & \textbf{Recall} & \textbf{F\_beta} & \textbf{Matthews} \\ \midrule
W2V Wiki & 0.6966 & 0.2211 & 0.5324 & 0.2649 & 0.2683 \\
W2V Abstracts 100D Cbow & 0.6327 & 0.1632 & 0.4567 & 0.2163 & 0.2203 \\
W2V Abstracts 200D Cbow & 0.6576 & 0.1738 & 0.4983 & 0.2340 & 0.2395 \\
W2V Abstracts 100D Skipgram & 0.6189 & 0.1880 & 0.4341 & 0.2293 & 0.2097 \\
W2V Abstracts 200D Skipgram & 0.6253 & 0.1900 & 0.4526 & 0.2300 & 0.2158 \\
FT Wiki & 0.3774 & 0.0772 & 0.3252 & 0.1161 & 0.1060 \\
FT Abstracts 100D Cbow & 0.5915 & 0.2322 & 0.5796 & 0.2879 & 0.2285 \\
FT Abstracts 200D Cbow & 0.6835 & 0.2480 & 0.6047 & 0.3120 & 0.2745 \\
FT Abstracts 100D Skipgram & 0.6391 & 0.2390 & 0.5792 & 0.2950 & 0.2467 \\
FT Abstracts 200D Skipgram & 0.6684 & 0.2440 & 0.5950 & 0.3070 & 0.2650 \\
BERT & 0.7737 & 0.4450 & 0.6913 & 0.4911 & 0.4643 \\
RoBERTa & 0.7047 & 0.3911 & 0.7150 & 0.4617 & 0.4072 \\
GPT-2 & 0.6461 & 0.3381 & 0.6230 & 0.3816 & 0.3463 \\
\textbf{SciBERT} & \textbf{0.8439} & \textbf{0.4895} & \textbf{0.7598} & \textbf{0.5399} & \textbf{0.5435} \\          \bottomrule
\end{tabular}
\caption{Results of Named Entity Recognition on ChemDNER Dataset}
\label{tab:chemdner_results}
\end{table}
\begin{table}[h!]
\centering

\begin{tabular}{@{}llllll@{}}
\toprule
\textbf{Model Name} & \textbf{Accuracy} & \textbf{Precision} & \textbf{Recall} & \textbf{F\_beta} & \textbf{Matthews}  \\ \midrule
W2V Wiki & 0.1866 & 0.2171 & 0.3203 & 0.2294 & 0.1514 \\
W2V Abstracts 100D Cbow & 0.2283 & 0.1863 & 0.3698 & 0.2279 & 0.1706 \\
W2V Abstracts 200D Cbow & 0.2456 & 0.1957 & 0.4205 & 0.2556 & 0.1967 \\
W2V Abstracts 100D Skipgram & 0.1245 & 0.1308 & 0.2874 & 0.1686 & 0.1031 \\
W2V Abstracts 200D Skipgram & 0.1650 & 0.1562 & 0.3162 & 0.1925 & 0.1314 \\
FT Wiki & 0.0562 & 0.0486 & 0.1715 & 0.0534 & 0.0259 \\
FT Abstracts 100D Cbow & 0.2223 & 0.1972 & 0.3307 & 0.2279 & 0.1596 \\
FT Abstracts 200D Cbow & 0.1916 & 0.2084 & 0.3621 & 0.2432 & 0.1644 \\
FT Abstracts 100D Skipgram & 0.1677 & 0.1995 & 0.3249 & 0.2259 & 0.1381 \\
FT Abstracts 200D Skipgram & 0.1929 & 0.2103 & 0.3619 & 0.2480 & 0.1621 \\
BERT & 0.2849 & 0.3158 & 0.4389 & 0.3416 & 0.2381 \\
RoBERTa & 0.2928 & 0.3059 & 0.4371 & 0.3313 & 0.2267 \\
GPT-2 & 0.0759 & 0.1466 & 0.2041 & 0.1412 & 0.0681 \\
\textbf{SciBERT} & \textbf{0.3538} & \textbf{0.3227} & \textbf{0.5384} & \textbf{0.3841} & \textbf{0.2985} \\ \bottomrule
\end{tabular}
\caption{Results of Named Entity Recognition on SciERC Dataset}
\label{tab:scierc_results}
\end{table}
\newpage

\subsubsection{Document Classification}
\begin{table}[h!]
\centering

\begin{tabular}{@{}llllll@{}}
\toprule
\textbf{Model Name}                   & \textbf{Accuracy} & \textbf{Precision} & \textbf{Recall} & \textbf{F\_beta}  & \textbf{Matthews} \\ \midrule
W2V Wiki & 0.8960 & 0.8979 & 0.8960 & 0.8940 & 0.8484 \\
W2V Abstracts 100D Cbow & 0.8891 & 0.8877 & 0.8891 & 0.8858 & 0.8383 \\
W2V Abstracts 200D Cbow & 0.9030 & 0.9027 & 0.9030 & 0.9008 & 0.8590 \\
W2V Abstracts 100D Skipgram & 0.9099 & 0.9097 & 0.9099 & 0.9076 & 0.8690 \\
\textbf{W2V Abstracts 200D Skipgram} & \textbf{0.9257} & \textbf{0.9254} & \textbf{0.9257} & \textbf{0.9243} & \textbf{0.8922} \\
FT Wiki & 0.9178 & 0.9196 & 0.9178 & 0.9167 & 0.8807 \\
FT Abstracts 100D Cbow & 0.8634 & 0.8683 & 0.8634 & 0.8599 & 0.8000 \\
FT Abstracts 200D Cbow & 0.8614 & 0.8660 & 0.8614 & 0.8591 & 0.7976 \\
FT Abstracts 100D Skipgram & 0.8574 & 0.8612 & 0.8574 & 0.8542 & 0.7910 \\
FT Abstracts 200D Skipgram & 0.8683 & 0.8725 & 0.8683 & 0.8658 & 0.8078 \\
BERT & 0.9198 & 0.9251 & 0.9198 & 0.9197 & 0.8838 \\
RoBERTa & 0.8376 & 0.8353 & 0.8376 & 0.8338 & 0.7622 \\
GPT-2 & 0.7366 & 0.7349 & 0.7366 & 0.7228 & 0.6057 \\
SciBERT & 0.8812 & 0.8831 & 0.8812 & 0.8768 & 0.8261 \\ \bottomrule
\end{tabular}
\caption{Results on Reuters 12 Dataset}
\label{tab:reuters12_results}
\end{table}
\begin{table}[h!]
\centering

\begin{tabular}{@{}llllll@{}}
\toprule
\textbf{Model Name}                   & \textbf{Accuracy} & \textbf{Precision} & \textbf{Recall} & \textbf{F\_beta}  & \textbf{Matthews} \\ \midrule
W2V Wiki & 0.8213 & 0.8196 & 0.8213 & 0.8186 & 0.7962 \\
W2V Abstracts 100D Cbow & 0.8750 & 0.8734 & 0.8750 & 0.8733 & 0.8574 \\
W2V Abstracts 200D Cbow & 0.8838 & 0.8826 & 0.8838 & 0.8823 & 0.8675 \\
W2V Abstracts 100D Skipgram & 0.8982 & 0.8972 & 0.8982 & 0.8970 & 0.8838 \\
\textbf{W2V Abstracts 200D Skipgram} & \textbf{0.9025} & \textbf{0.9015} & \textbf{0.9025} & \textbf{0.9015} & \textbf{0.8888} \\
FT Wiki & 0.8758 & 0.8744 & 0.8758 & 0.8745 & 0.8583 \\
FT Abstracts 100D Cbow & 0.8522 & 0.8503 & 0.8522 & 0.8504 & 0.8313 \\
FT Abstracts 200D Cbow & 0.8564 & 0.8545 & 0.8564 & 0.8545 & 0.8362 \\
FT Abstracts 100D Skipgram & 0.8499 & 0.8477 & 0.8499 & 0.8479 & 0.8287 \\
FT Abstracts 200D Skipgram & 0.8569 & 0.8548 & 0.8569 & 0.8549 & 0.8366 \\
BERT & 0.8318 & 0.8288 & 0.8318 & 0.8293 & 0.8080 \\
RoBERTa & 0.8113 & 0.8092 & 0.8113 & 0.8085 & 0.7847 \\
GPT-2 & 0.6735 & 0.6690 & 0.6735 & 0.6677 & 0.6277 \\
SciBERT & 0.8445 & 0.8427 & 0.8445 & 0.8423 & 0.8226 \\
\bottomrule
\end{tabular}
\caption{Results on BioChem 8 Dataset}
\label{tab:biochem_results}
\end{table}
\newpage

\subsubsection{Semantic Partitioning}
\begin{table}[h!]
\centering

\begin{tabular}{@{}ll@{}}
\toprule
\textbf{Model Name}                   & \textbf{Average Score} \\ \midrule
\textbf{W2V Wiki} & \textbf{0.5679} \\
W2V Abstracts 100D Cbow & 0.1949 \\
W2V Abstracts 200D Cbow & 0.1850 \\
W2V Abstracts 100D Skipgram & 0.3681 \\
W2V Abstracts 200D Skipgram & 0.3184 \\
FT Wiki & 0.5089 \\
FT Abstracts 100D Cbow & 0.4003 \\
FT Abstracts 200D Cbow & 0.3864 \\
FT Abstracts 100D Skipgram & 0.4000 \\
FT Abstracts 200D Skipgram & 0.3828 \\
BERT & 0.2437 \\
RoBERTa & 0.3535 \\
GPT-2 & 0.1075 \\
SciBERT & 0.3874 \\
\bottomrule
\end{tabular}
\caption{Results of Semantc Partitioning on Conll 2003 Dataset}
\label{tab:conll_sp_results}
\end{table}
\begin{table}[h!]
\centering

\begin{tabular}{@{}ll@{}}
\toprule
\textbf{Model Name}                   & \textbf{Average Score} \\ \midrule
W2V Wiki & 0.6318 \\
W2V Abstracts 100D Cbow & \textbf{0.9234} \\
\textbf{W2V Abstracts 200D Cbow} & \textbf{0.9234} \\
W2V Abstracts 100D Skipgram & 0.3850 \\
W2V Abstracts 200D Skipgram & 0.3850 \\
FT Wiki & -0.0204 \\
FT Abstracts 100D Cbow & 0.6318 \\
FT Abstracts 200D Cbow & 0.3850 \\
FT Abstracts 100D Skipgram & 0.4505 \\
FT Abstracts 200D Skipgram & 0.3850 \\
BERT & 0.0542 \\
RoBERTa & 0.5070 \\
GPT-2 & 0.0081 \\
SciBERT & 0.6968 \\
\bottomrule
\end{tabular}
\caption{Results of Semantc Partitioning on AM Patent Dataset}
\label{tab:ampatent_results}
\end{table}
\newpage

\subsubsection{Tokenization}
\label{sec:tok_results}
\begin{table}[h!]
\centering
\noindent\makebox[\textwidth]{
\begin{tabular}{@{}llllll@{}}
\toprule
\textbf{Embedding   Model} &
  \textbf{Algorithm} &
  \textbf{Tokenization Method} &
  \multicolumn{1}{c}{\textbf{SemEval}} &
  \multicolumn{1}{c}{\textbf{STS 18}} &
  \multicolumn{1}{c}{\textbf{STS 19}} \\ \midrule
\textbf{Word2Vec} & Cbow     & -         & 0.5967          & 0.7349          & 0.5975          \\
\textbf{Word2Vec} & Skipgram & -         & \textbf{0.6567} & \textbf{0.7514} & \textbf{0.6300} \\
\textbf{}         &          &           & \textbf{}       & \textbf{}       & \textbf{}       \\
\textbf{Word2Vec} & Cbow     & BPE       & 0.4313          & 0.6217          & 0.5004          \\
\textbf{Word2Vec} & Cbow     & Unigram   & 0.4567          & 0.6279          & 0.5221          \\
\textbf{Word2Vec} & Cbow     & WordPiece & 0.4506          & 0.5429          & 0.4607          \\
\textbf{}         &          &           &                 &                 &                 \\
\textbf{Word2Vec} & Skipgram & BPE       & 0.5469          & 0.6408          & 0.5443          \\
\textbf{Word2Vec} & Skipgram & Unigram   & 0.5904          & 0.6616          & 0.5706          \\
\textbf{Word2Vec} & Skipgram & WordPiece & 0.5378          & 0.6458          & 0.5688          \\
\textbf{}         &          &           &                 &                 &                 \\
\textbf{FastText} & Cbow     & N-grams   & 0.4300          & 0.4382          & 0.4031          \\
\textbf{FastText} & Skipgram & N-grams   & 0.4310          & 0.4363          & 0.4056          \\ \bottomrule
\end{tabular}}
\caption{Tokenization Results on Sentence Similarity Datasets. Metric: Pearson Correlation}
\label{tab:tok_ss}
\end{table}
\begin{table}[h!]
\centering
\noindent\makebox[\textwidth]{
\begin{tabular}{@{}llllll@{}}
\toprule
\textbf{Embedding   Model} &
  \textbf{Algorithm} &
  \textbf{Tokenization Method} &
  \multicolumn{1}{c}{\textbf{Conll2003}} &
  \multicolumn{1}{c}{\textbf{ChemdNER}} &
  \multicolumn{1}{c}{\textbf{SciERC}} \\ \midrule
\textbf{Word2Vec} & Cbow     & -         & 0.3973 & 0.2163 & 0.2279 \\
\textbf{Word2Vec} & Skipgram & -         & 0.4040 & 0.2293 & 0.1686 \\
\textbf{}         &          &           &        &        &        \\
\textbf{Word2Vec} & Cbow     & BPE       & 0.5024 & 0.3402 & 0.3117 \\
\textbf{Word2Vec} & Cbow     & Unigram   & 0.4693 & 0.3109 & 0.2992 \\
\textbf{Word2Vec} &
  Cbow &
  WordPiece &
  \textbf{0.5353} &
  \textbf{0.3429} &
  \textbf{0.3254} \\
\textbf{}         &          &           &        &        &        \\
\textbf{Word2Vec} & Skipgram & BPE       & 0.3937 & 0.2611 & 0.1963 \\
\textbf{Word2Vec} & Skipgram & Unigram   & 0.3904 & 0.2450 & 0.1978 \\
\textbf{Word2Vec} & Skipgram & WordPiece & 0.4366 & 0.2725 & 0.1997 \\
\textbf{}         &          &           &        &        &        \\
\textbf{FastText} & Cbow     & N-grams   & 0.4097 & 0.2879 & 0.4031 \\
\textbf{FastText} & Skipgram & N-grams   & 0.4238 & 0.2950 & 0.4056 \\ \bottomrule
\end{tabular}}
\caption{Tokenization Results on Named Entity Recognition. Metric: F-Beta}
\label{tab:tok_ner}
\end{table}
\begin{table}[h!]
\centering
\noindent\makebox[\textwidth]{
\begin{tabular}{@{}lllll@{}}
\toprule
\textbf{Embedding   Model} &
  \textbf{Algorithm} &
  \textbf{Tokenization Method} &
  \multicolumn{1}{c}{\textbf{Reuters 12}} &
  \multicolumn{1}{c}{\textbf{Core BioChem 8}} \\ \midrule
\textbf{Word2Vec} & Cbow     & -         & 0.8858          & 0.8733          \\
\textbf{Word2Vec} & Skipgram & -         & \textbf{0.9099} & \textbf{0.8970} \\
\textbf{}         &          &           & \textbf{}       & \textbf{}       \\
\textbf{Word2Vec} & Cbow     & BPE       & 0.8427          & 0.8515          \\
\textbf{Word2Vec} & Cbow     & Unigram   & 0.8287          & 0.8430          \\
\textbf{Word2Vec} & Cbow     & WordPiece & 0.8347          & 0.8425          \\
\textbf{}         &          &           &                 &                 \\
\textbf{Word2Vec} & Skipgram & BPE       & 0.8669          & 0.8784          \\
\textbf{Word2Vec} & Skipgram & Unigram   & 0.8695          & 0.8813          \\
\textbf{Word2Vec} & Skipgram & WordPiece & 0.8492          & 0.8702          \\
\textbf{}         &          &           &                 &                 \\
\textbf{FastText} & Cbow     & N-grams   & 0.8599          & 0.8545          \\
\textbf{FastText} & Skipgram & N-grams   & 0.8542          & 0.8479         \\ \bottomrule
\end{tabular}}
\caption{Tokenization Results on Document Classification Datasets. Metric: F-Beta}
\label{tab:tok_dc}
\end{table}
\begin{table}[h!]
\centering
\noindent\makebox[\textwidth]{
\begin{tabular}{@{}lllll@{}}
\toprule
\textbf{Embedding   Model} &
  \textbf{Algorithm} &
  \textbf{Tokenization Method} &
  \multicolumn{1}{c}{\textbf{Conll 2003}} &
  \multicolumn{1}{c}{\textbf{AM Patent}} \\ \midrule
\textbf{Word2Vec} & Cbow     & -         & 0.1949 & \textbf{0.9234} \\
\textbf{Word2Vec} & Skipgram & -         & 0.3681 & 0.3850          \\
\textbf{}         &          &           &        &                 \\
\textbf{Word2Vec} & Cbow     & BPE       & 0.2791 & 0.1653          \\
\textbf{Word2Vec} & Cbow     & Unigram   & 0.1743 & 0.1639          \\
\textbf{Word2Vec} & Cbow     & WordPiece & 0.2790 & 0.6500          \\
\textbf{}         &          &           &        &                 \\
\textbf{Word2Vec} & Skipgram & BPE       & 0.2689 & 0.0989          \\
\textbf{Word2Vec} & Skipgram & Unigram   & 0.2222 & 0.0740          \\
\textbf{Word2Vec} & Skipgram & WordPiece & 0.2965 & 0.0291          \\
\textbf{}         &          &           &        &                 \\
\textbf{FastText} & Cbow     & N-grams   & \textbf{0.4003} & 0.6318          \\
\textbf{FastText} & Skipgram & N-grams   & 0.4000 & 0.4505         \\ \bottomrule
\end{tabular}}
\caption{Tokenization Results on Semantic Partitioning Datasets. Metric: Average Score}
\label{tab:tok_sp}
\end{table}
\newpage

\end{document}